\documentclass[10pt,twocolumn,letterpaper]{article}

\usepackage{cvpr}              % To produce the CAMERA-READY version
\usepackage{graphicx}
\usepackage{amsmath}
\usepackage{amssymb}
\usepackage{booktabs}

\usepackage{multirow}
\usepackage{makecell}
\usepackage{appendix}
\usepackage[pagebackref,breaklinks,colorlinks]{hyperref}

\usepackage[capitalize]{cleveref}
\crefname{section}{Sec.}{Secs.}
\Crefname{section}{Section}{Sections}
\Crefname{table}{Table}{Tables}
\crefname{table}{Tab.}{Tabs.}

\def\cvprPaperID{1526} % *** Enter the CVPR Paper ID here
\def\confName{CVPR}
\def\confYear{2023}

\begin{document}

%\title{Patch-wise Relational Knowledge Distillation in Compressing GANs}
\title{Exploring Content Relationships for Distilling Efficient GANs}

\author{Lizhou You$^{1}$, Mingbao Lin$^3$, Tie Hu$^1$, Fei Chao$^{1}$, Rongrong Ji$^{1,2}$\thanks{Corresponding Author}\\
$^1$MAC Lab, School of Informatics, Xiamen University \\
$^2$Institute of Artificial Intelligence, Xiamen University\\
$^3$Tencent Youtu Lab\\
{\tt\small \{youlizhou, lmbxmu, hutie\}@stu.xmu.edu.cn, \{fchao, rrji\}@xmu.edu.cn}
}

\maketitle

%%%%%%%%% ABSTRACT
\begin{abstract}
This paper proposes a content relationship distillation (CRD) to tackle the over-parameterized generative adversarial networks (GANs) for the serviceability in cutting-edge devices. In contrast to traditional instance-level distillation, we design a novel GAN compression oriented knowledge by slicing the contents of teacher outputs into multiple fine-grained granularities, such as row/column strips (global information) and image patches (local information), modeling the relationships among them, such as pairwise distance and triplet-wise angle, and encouraging the student to capture these relationships within its output contents. 
Built upon our proposed content-level distillation, we also deploy an online teacher discriminator, which keeps updating when co-trained with the teacher generator and keeps freezing when co-trained with the student generator for better adversarial training.
We perform extensive experiments on three benchmark datasets, the results of which show that our CRD reaches the most complexity reduction on GANs while obtaining the best performance in comparison with existing methods. For example, we reduce MACs of CycleGAN by around 40$\times$ and parameters by over 80$\times$, meanwhile, 46.61 FIDs are obtained compared with these of 51.92 for the current state-of-the-art. Code of this project is available at \url{https://github.com/TheKernelZ/CRD}.
% 替换链接后pdf单行只有一个单词，最后一句的描述稍微进行了改变，原文如下:
% Our project is publicly available at \url{https://github.com/TheKernelZ/CRD}.
\end{abstract}

%%%%%%%%% BODY TEXT
\section{Introduction}
\label{sec:intro}

Generative adversarial networks (GANs)~\cite{goodfellow2014generative}, consisting of a generator and a discriminator, attempt to find a balance between them to make outputs of the generator alike to real images.
Recent years have witnessed the power of GANs in leading the substantial development of image generation tasks such as image synthesis~\cite{karras2019style, karras2020analyzing, brock2018large, radford2015unsupervised, zhang2019self}, style transfer~\cite{gatys2016image, gatys2015texture, xu2021drb}, image-to-image translation~\cite{isola2017image, zhu2017unpaired, chen2018cartoongan, choi2018stargan}, \emph{etc}.
Different from traditional vision tasks, such as classification, image generation task seems harder and suffers from heavier burden of computations and parameters, raising a great challenge for GANs deployment on resource-inadequate devices.
Both academia and industry have called for economical GANs for a long time. 
As a result, GAN compression has become one of the most valuable topics currently. Typical methods resort to network pruning~\cite{li2021revisiting, li2022learning, chen2021gans}, neural architecture search (NAS)~\cite{fu2020autogan, jin2021teachers, li2020gan}, weight quantization~\cite{wang2020gan, wang2019qgan},  knowledge distillation~\cite{ren2021online, li2020gan, hou2021slimmable, aguinaldo2019compressing, wang2018kdgan, chen2020distilling, li2020semantic}, \emph{etc}.

\begin{figure}[!t]
  \centering
  \includegraphics[width=0.46\textwidth]{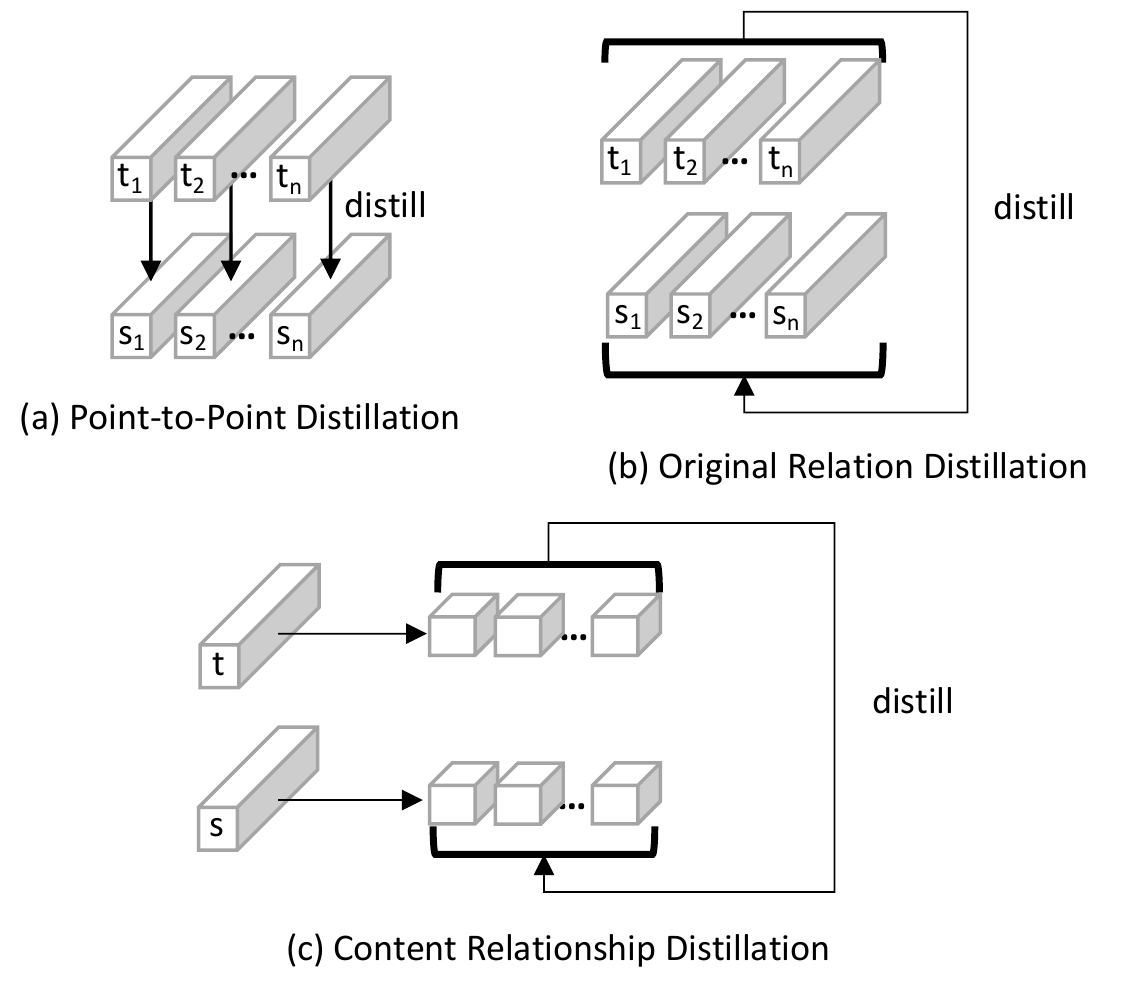}
\vspace{-0.5em}
  \caption{Instance-level distillation includes (a) vanilla point-by-point knowledge transferring~\cite{hinton2015distilling} and (b) relational knowledge transferring~\cite{park2019relational}. (c) Our content-level distillation excavates content structure relationships between teacher and student.}
  \label{three distillation methods}
  \vspace{-1.0em}
\end{figure}

\begin{figure*}[!t]
  \centering
  \includegraphics[width=0.9\textwidth]{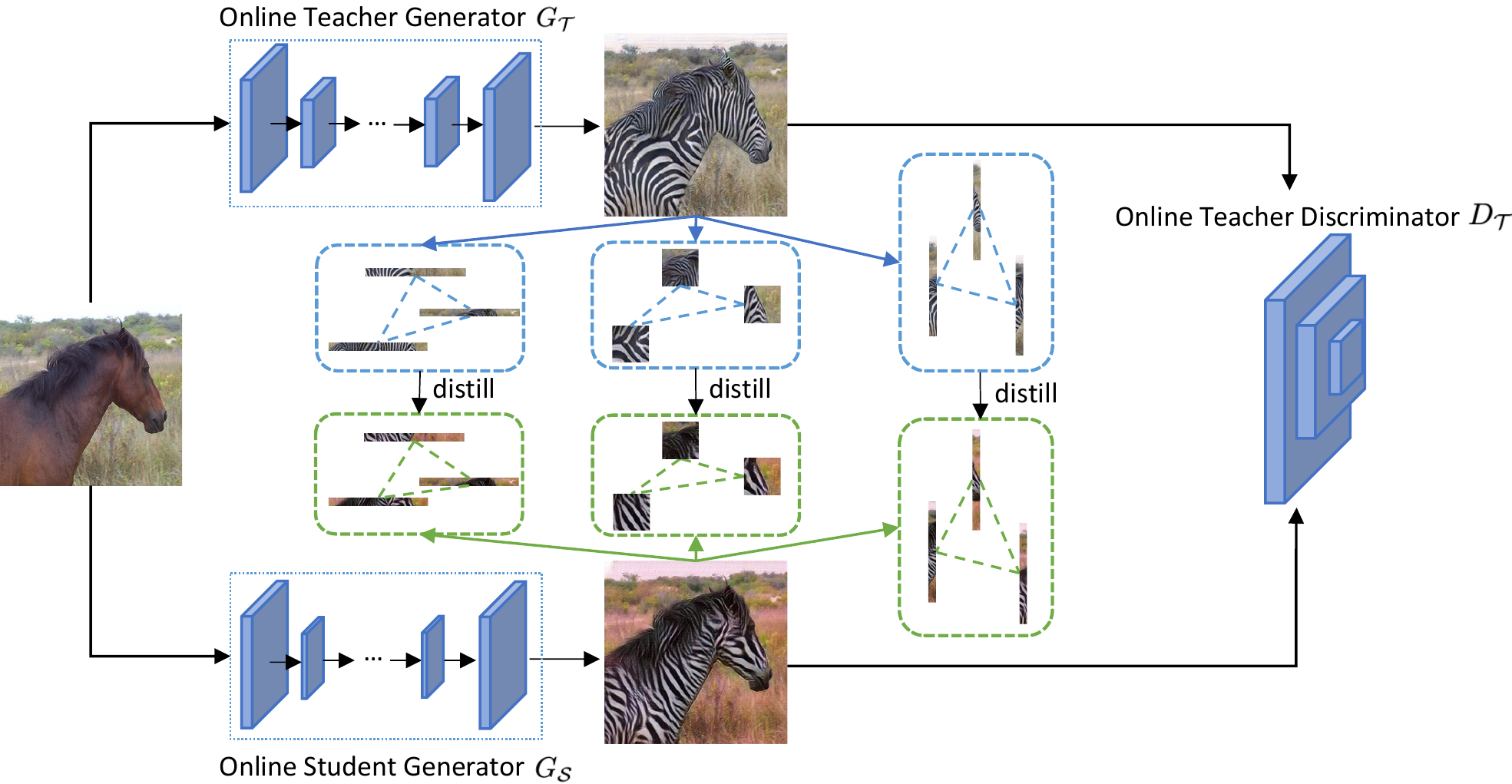}
\vspace{-0.5em}
\caption{Our CRD framework. We split the output images into row/column strips and image patches, and minimize their pairwise distance and triplet-wise angle structures between the students/teachers. Teacher discriminator is online updated when co-trained with the teacher generator and frozen when co-trained with the student generator.}
\vspace{-1.0em}
  \label{PRKD}
\end{figure*}

Among all of these methodologies, knowledge distillation (KD) is demonstrated to be particularly efficacious in contemporary GAN compression and has almost become a standard complement to other compression methods.
For example, CAT~\cite{jin2021teachers} introduces IncResBlock in origin GANs, the results of which serve as a teacher to guide student learning. The kernel sizes in IncResBlock are determined in a NAS fashion.
SlimGANs~\cite{hou2021slimmable} presents slimmable pruning~\cite{yu2018slimmable} to GAN compression, which divides a GANs model into four weight-shared subnetworks and the larger individuals are adopted to distill the smaller ones.
GAN-Slimming~\cite{wang2020gan} migrates the conventional pipelines of pruning, quantization, and coding in Deep Compression~\cite{han2015deep}. Differently, the coding progress is replaced with distillation for a better effect on compressing GANs.
Albeit the extensive efforts, earlier methods, as we go over in this paper, endure two limitations that prevent further development of GAN compression.

First, most knowledge distillation implementations of GAN compression are extension of conventional KD methods~\cite{hinton2015distilling, romero2014fitnets, han2015deep} for vision tasks, such as classification and detection, where the teacher outputs are point-by-point transferred to the student model, as illustrated in Fig.\,\ref{three distillation methods}(a). However, GANs are famous for its image generation ability which is strict in structural relationships. For example, when transferring a horse image to a zebra image, we expect the distance structure between ``eye’’ and ``noise’’ in the horse can be well retained in the zebra image. Though recent work~\cite{park2019relational} introduces relational knowledge distillation to transfer structure-wise relations, as manifested in Fig.\,\ref{three distillation methods}(b), a more profound attempt on GAN distillation remains unexplored.
Second, most previous KD methods concentrate on excavating instance-level information, such as distribution logits and feature salience, from an entire image instance. Such a paradigm, as we analyze in this paper, might not be the most favourable in GAN distillation. Instance-level distillation benefits the student model to form more robust high-level image features, which are primarily helpful for tasks like classification and detection. On the contrary, the generative task pays more attention to the image contents, in which more fine-grained content-level knowledge is more valuable.

Above all, excavating content knowledge and relationship structure remains unexplored in GANs oriented distillation, and might be a promise of boosting the compressed GANs performance. 
We believe the above two issues are interrelated and can be addressed in a unified framework. Accordingly, we propose Content Relationship Distillation (abbreviated as CRD) in this paper for better learning an efficient GANs model, as outlined in Fig.\,\ref{three distillation methods}(c).
The main innovations in our CRD are two-fold including modeling content-level structural relationships for knowledge transferring in Sec.\,\ref{Patch-wise Relational Knowledge Distillation} and adversarially training teacher discriminator in an online updating-freezing manner in Sec.\,\ref{online nash}. A specific framework of our CRD can be referred to Fig.\,\ref{PRKD}. In the rear of generators, we seize the global and local content information by slicing an output image into row/column strips and image patches, results of which act as our basic granularity for building relationships such as pairwise distance structure and triplet-wise angle structure. Between the teacher and student outputs, we encourage the content structure to be alike which can be well supported by off-the-shelf algorithm~\cite{park2019relational} stemming from prototypical instance-level distillation. 
On the basis of fine-grained content distillation, we throw away a well-pretrained discriminator which is too strong for the student generator to maintain the Nash equilibrium~\cite{nash1951non}. On the contrary, we train the teacher discriminator from scratch in an online updating-freezing fashion in which discriminator weights are updated when co-trained with the teacher generator while frozen when co-trained with the student generator. In this fashion, a better Nash equilibrium is observed in adversarially training GANs.
We have performed extensive experiments on three acknowledged benchmark datasets~\cite{zhu2017unpaired, yu2014fine} to demonstrate the efficacy of our CRD in boosting the performance of compressed GANs. Experimental results indicate that our CRD not only owns the most GANs complexity reduction, but also leads to the best performance. For example, we reduce MACs of CycleGAN~\cite{zhu2017unpaired} by around 40$\times$ and parameters by over 80$\times$ on horse$\rightarrow$zebra dataset~\cite{zhu2017unpaired}. Meanwhile, 46.61 FIDs are obtained compared with these of 51.92 for the current state-of-the-art (SoTA) ~\cite{ren2021online}.

This work addresses the problem of distilling a light-weight student generator to enable practical deployment of GANs models on resource-inadequate platforms. The key contributions of this paper include: 
(1) One novel GANs oriented content relationship distillation method.
(2) Online updating-freezing discriminator learning for better Nash equilibrium in GAN compression. 
(3) Significant performance increase as well as model complexity decrease.

%
%The remainder of this paper is outlined in this following:
%
%We briefly revisit some related studies in Sec.\,\ref{sec:related}. Then, we present details of our CRD for GAN distillation in Sec.\,\ref{sec:methodology}. In Sec.\,\ref{sec:experimentation}, discussions on the performance of our method in comparison with off-the-shelf methods~\cite{zhu2017unpaired, ren2021online, li2020gan, shu2019co, wang2020gan, fu2020autogan, zhang2022wavelet, jung2022exploring, li2022learning, jin2021teachers} are manifested. Moreover, a brief discussion on the limitations of this work is given in Sec.\,\ref{sec:limitations}, towards some directions for further excavation of our CRD. Finally, we give a conclusion in Sec.\,\ref{sec:conclusion}.

%------------------------------------------------------------------------
\section{Related Work}
\label{sec:related}

\textbf{Generative Adversarial Networks}.
Generative adversarial networks (GANs)~\cite{goodfellow2014generative} have substantially boosted performance of various generative tasks. DCGAN~\cite{radford2015unsupervised}, one of the most influential improvements on GANs structure, introduces deep convolutional networks to enhance the generative adversarial capabilities. And then, recent years have witnessed the great efforts on improving network losses and network structures~\cite{isola2017image, zhu2017unpaired, choi2018stargan, karras2019style, karras2020analyzing}.
Among the majorities of generative tasks, image-to-image translation intends to translate an image from the source domain to the target domain, leading to the increasing popularity of GANs applications. CycleGAN~\cite{zhu2017unpaired} for the first time implements unpaired image-to-image translation by introducing a cycle-consistency loss. Pix2Pix~\cite{isola2017image} succeeds in supervised settings due mostly to its ability for balancing performance and consumption in paired image-to-image tasks. 
Despite the progress, the performance increase also stems from rising computation and memory costs, which are unacceptable for cutting-edge devices in which the available resources are very limited. In this paper, we focus on developing an efficient model by cutting off the redundancy in existing GANs.

\textbf{Knowledge Distillation}.
Knowledge distillation (KD) transfers dark knowledge from a high-capacity teacher network to student.
The pioneering work dates back to~\cite{hinton2015distilling} which minimizes $\ell_2$ distance of output logits between the students and the teachers.
DKD~\cite{zhao2022decoupled} decouples logit distillation into weighted target classes and non-target classes.
In addition to output logits~\cite{zhao2022decoupled, hinton2015distilling}, other formats of knowledge are excavated as well such as intermediate feature maps~\cite{romero2014fitnets, chen2021distilling, heo2019comprehensive, tian2019contrastive, chen2021wasserstein}.
FitNets~\cite{romero2014fitnets} aligns intermediate features between students and teachers by 1$\times$1 convolution first before minimizing their distance.
These conventional KD methods transfer point-by-point outputs from a teacher model to a student model. In contrast, relational knowledge distillation (RKD)~\cite{park2019relational} was proposed to transfer the relation of the output structure-wise, such as distance and angle as shown in Fig.\,\ref{three distillation methods}.
Nevertheless, existing methods all focus on an instance-level distillation where information hints across entire image instances are extracted.
In this paper, our content relationship distillation (CRD) also models structural relationships supported by RKD~\cite{park2019relational}. More efforts are made to excavate GANs-oriented content-level relationships to learn a better student model.

\textbf{GAN Compression}.
Though GANs excel at synthesizing realistic images, they are also notoriously computational intensive~\cite{li2022learning}. Therefore, compressing GANs has received substantial attention due mostly to the urgent applications on resource-inadequate devices.
In addition to KD, typical compression methodologies include weights pruning~\cite{shu2019co, chen2021gans}, network quantization~\cite{wang2020gan, wang2019qgan}, neural architecture search~\cite{li2020gan, fu2020autogan, jin2021teachers}, \emph{etc}.
Shu \emph{et al}.~\cite{shu2019co} for the first time developed a co-evolutionary algorithm to prune CycleGAN~\cite{zhu2017unpaired}. They encoded generators of two image domains into populations and investigated important convolution filters in a synergistic optimal fashion.
Alike to~\cite{han2015deep}, GAN-Slimming~\cite{wang2020gan} reduces complexity of GANs by sequentially pruning, quantization and distillation.
GAN Compression~\cite{li2020gan} trains a ``Once-for-all'' student generator using feature distillation and neural architecture search.
OMGD~\cite{ren2021online}, which conducts discriminator-free distillation, introduces teachers in wider and deeper network structures, intermediate features and output logits are then distilled to the students.
GCC~\cite{li2021revisiting} revisits the efficacy of the discriminator in GAN compression and introduces a generator-discriminator cooperative compression scheme where the discriminator is pruned as well to achieve Nash equilibrium.
Motivated by GCC, we also consider maintaining Nash equilibrium in GANs training. Differently, we do not conduct any professional pruning surgery upon discriminator, but propose to cooperate generators with teacher discriminator in an online updating-freezing manner.

\section{Methodology}\label{sec:methodology}

\subsection{Background} \label{background}
Considering a teacher GAN model $\mathcal{T}$ and a student GAN model $\mathcal{S}$, we denote $f_\mathcal{T}$ and $f_\mathcal{S}$ as functions of the teacher $\mathcal{T}$ and the student $\mathcal{S}$. Notice $f$ can be the output of any layer of the network in GANs. Given an input $x$, traditional knowledge distillation minimizes the point-by-point difference of outputs $f_\mathcal{T}(x)$ and $f_\mathcal{S}(x)$ as $\ell\big(f_\mathcal{T}(x), f_\mathcal{S}(x)\big)$ in which the penalty function $\ell(\cdot,\cdot)$ minimizes the difference.

In contrast, we model the structural relationships in an input sequence $\mathcal{X}^N = \{(x_1, x_2, ..., x_N) | x_1 \neq x_2 \neq ... \neq x_N\}$. Our implementation procedure primarily stems from the relational knowledge distillation~\cite{park2019relational} that formulates the learning process as:
\begin{equation}\label{rkd}
    \mathcal{L}_{RKD} = \sum_{x \in \mathcal{X}^N} \ell\Big( \phi\big(f_{\mathcal{T}}(x)\big), \phi\big(f_{\mathcal{S}}(x)\big) \Big),
\end{equation}
where $\phi(\cdot)$ models the structural relationships among its inputs, such as pairwise distance in $\mathcal{X}^2$ and triplet-wise angle in $\mathcal{X}^3$ that we briefly revisit in the following.

\textbf{Pairwise Distance Distillation}. Considering $\mathcal{X}^2 = \{(x_i, x_j) | i \neq j\}$, for any $x \in \mathcal{X}^2$, $\phi(\cdot)$ can be defined as the Euclidean distance between elements in the input tuple:
\begin{equation}
    \phi_d(x) = \frac{1}{\mu}\| f(x_i) - f(x_j) \|_2,
\end{equation}
which empowers the students to learn the relation of distance structure from the teacher outputs. The $\mu = (\sum_{x \in \mathcal{X}^2} ||f(x_i) - f(x_j)||_2)/N$ is a normalization factor. Accordingly, Eq.\,(\ref{rkd}) can be explicitly rewritten as:
\begin{equation}\label{rkd-d}
   \mathcal{L}_{RKD-D} = \sum\limits_{x \in \mathcal{X}^2}  \ell\Big( \phi_d\big(f_{\mathcal{T}}(x)\big), \phi_d\big(f_{\mathcal{S}}(x)\big) \Big).
\end{equation}

\textbf{Triplet-wise Angle Distillation}. Considering $\mathcal{X}^3 = \{(x_i, x_j, x_k) | i \neq j \neq k\}$, for any $x \in \mathcal{X}^3$, $\phi(\cdot)$ model the angle structure between residues $e^{ij} = (x_i - x_j) / \|x_i - x_j\|_2$ and $e^{jk} = (x_j - x_k) / \|x_j - x_k\|_2$ as:
\begin{equation}
    \phi_a(x) = \cos{\angle{x_i x_j x_k}} = \langle e^{i j}, e^{j k} \rangle.
\end{equation}

In this case, Eq.\,(\ref{rkd}) is explicitly rewritten as:
\begin{equation}\label{rkd-a}
    \mathcal{L}_{RKD-A} = \sum\limits_{x \in \mathcal{X}^3} \ell\Big(\phi_a\big(f_\mathcal{T}(x)\big), \phi_a\big(f_\mathcal{S}(x)\big)\Big).
\end{equation}

As for the penalty function $\ell(\cdot)$, following~\cite{park2019relational}, we consider the Huber loss as:
\begin{equation} \label{huber loss}
    \ell(x, x') = \left\{
             \begin{array}{lr}
             \frac{1}{2}(x-x')^2 & |x-x'| \leq 1, \\
             |x-x'| - \frac{1}{2} & \text{otherwise}. \\
             \end{array}
\right.
\end{equation}

Our motivation of modeling structural relationships can be well implemented through off-the-shelf study~\cite{park2019relational} as described above. Therefore, we put our effort into exploring the specific format of $f(x)$, and propose GANs-oriented content relationships for learning an efficient student model with better performance.

\subsection{Content Relationship Distillation} \label{Patch-wise Relational Knowledge Distillation}
After looking at the GANs framework, we realize a huge gap between GAN distillation and traditional vision task distillation. The characteristic of vision tasks, such as classification and detection, mostly benefits from learning robust image features, therefore they urge more for distilling instance-level information such as distribution logits and feature outputs. Even the vanilla relational knowledge distillation~\cite{park2019relational} also concentrates efforts on preserving relations between image instances.
Very differently, GANs are mostly applied in generative tasks such as image synthesis, image-to-image translation and image editing, which pay attention to image contents. Therefore, the peculiarity of GANs requires to model content-level relationships for a better distillation.

To that effect, as illustrated in Fig.\,\ref{PRKD}, we discard the prototypical instance-level distillation. On the contrary, we utilize the fact that the generator of GANs results in an image outputs, upon which we design a unique content relationship distillation. Specifically, given an input image $x$, let $f(x) \in \mathbb{R}^{c \times h \times w}$ be the final image outputs of the generator where $c$, $h$, and $w$ respectively denote image channel, height, and width. 
We formally define an image splitting function $h(\cdot)$ as: 
\begin{equation}
    h\big(f(x)\big) \to \mathcal{C}, \mathcal{R}, \mathcal{P}.
\end{equation}

Herein, $\mathcal{C} = \{ c_i \}_{i=1}^w$ where $c_i \in \mathbb{R}^{c \times h}$ is the $i$-th column strip in $f(x)$; 
$\mathcal{R} = \{ r_i \}_{i=1}^h$ where $r_i \in \mathbb{R}^{c \times w}$ is the $i$-th row strip in $f(x)$.
As for $\mathcal{P}$, we first split an image $y$ into several non-overlapping patches of size $n \times m$. Then, $\mathcal{P} =  \{ p_i \}_{i=1}^{(hw)/(nm)}$ where $p_i \in \mathbb{R}^{c \times n \times m}$ stands for the $i$-th image patch in $f(x)$. %Notice we suppose $\mathcal{Y}$ is an ordered set here. 
As can be seen, we explore the content relationships of the generated images through smaller fine-grained granularities, including image columns, image rows, and image patches. The image patches $\mathcal{P}$ maintain local region information while the image columns $\mathcal{C}$ and image rows $\mathcal{R}$ consider long-range global information.

We form pairwise contents as $\mathcal{Y}^2 = \mathcal{C}^2 \cup \mathcal{R}^2 \cup \mathcal{P}^2$ where $\mathcal{C}^2 = \{ (c_i, c_j)| i \neq j\}$, $\mathcal{R}^2 = \{ (r_i, r_j)| i \neq j\}$, $\mathcal{P}^2 = \{ (p_i, p_j)| i \neq j \}$. Then, the pairwise distance distillation of Eq.\,(\ref{rkd-d}) in our content relationship situation becomes:
\begin{equation}\label{crdd}
%\begin{split}
\mathcal{L}_{CRD-D} = \sum_{\mbox{\tiny$\begin{array}{c}
t_i \in \mathcal{Y}^2_{\mathcal{T}}, 
s_i \in \mathcal{Y}^2_{\mathcal{S}}\end{array}$ }}\ell\big( \phi_d(t_i), \phi_d(s_i) \big).
%\end{split}
\end{equation}

Similarly, we have $\mathcal{Y}^3 = \mathcal{C}^3 \cup \mathcal{R}^3 \cup \mathcal{P}^3$ where $\mathcal{C}^3 = \{ (c_i, c_j, c_k)| i \neq j \neq k \}$, $\mathcal{R}^3 = \{ (r_i, r_j, r_k)| i \neq j \neq k \}$, $\mathcal{P}^3 = \{ (p_i, p_j, p_k)| i \neq j \neq k \}$. Also, the triplet-wise angle distillation of Eq.\,(\ref{rkd-a}) in our content relationship situation is reformulated in the following:
\begin{equation}\label{crda}
\begin{split}
\mathcal{L}_{CRD-A} = \sum_{\mbox{\tiny$\begin{array}{c}
t_i \in \mathcal{Y}^3_{\mathcal{T}},
s_i \in \mathcal{P}^3_{\mathcal{S}}\end{array}$ }}\ell\big( \phi_a(t_i), \phi_a(s_i) \big).
\end{split}
\end{equation}

Combining Eq.\,(\ref{crdd}) and Eq.\,(\ref{crda}) lead to our final content relationship distillation loss:
\begin{equation}
    \mathcal{L}_{CRD} = \mathcal{L}_{CRD-D} + \lambda_a \mathcal{L}_{CRD-A}
\end{equation}
and $\lambda_a$ is used to balance angle and distance losses.

\subsection{Online Teacher Discriminator} \label{online nash}
Traditional GANs consist of a generator $G$ and a discriminator $D$, two of which content with each other in the form of a zero-sum game. Given an input image $x$ as well as a reference image $y$, the generator $G$ intends to match its output distribution as closely as possible to the reference distribution, \emph{i.e.}, $G(x) \approx y$. The task of discriminator $D$ is to output a value close to 1 when its input is from the reference distribution, or 0 when its input is the generator distribution.
The generator $G$ and the discriminator $D$ are optimized in an alternating manner until converging to the Nash equilibrium, with the objective function as:
\begin{equation}
    \begin{split}
        \min\limits_{G} \max\limits_{D} \mathcal{L}(D, G) = \mathbb{E}_{y \sim p_{data}} \log D(y) \\ + \mathbb{E}_{x \sim p(x)}\bigg[\log \Big(1-D\big(G(x)\big)\Big)\bigg],
    \end{split}
\end{equation}
where the generator aims to minimize the objective, and the discriminator aims to maximize the objective.

Note that Nash equilibrium is reached when the generator and discriminator are of similar capacities. Previous methods~\cite{chen2020distilling, lin2021anycost, jin2021teachers} deploy a well pre-trained discriminator, which however is too strong for the from-scratch-training generator such that the resulting model is often stuck in a local optimum. Ren~\emph{et al}.~\cite{ren2021online} chose to train discriminator from scratch when training teacher generator while discarding it when training student generator.
Given that our content relationship distillation pulls up the capacity of the student generator to align with that of the teacher generator, we realize an online teacher discriminator can well assist the student generator as well. Therefore, we apply a from-scratch-training teacher discriminator to adversarially train with the teacher generator and the student generator simultaneously. Differently, the discriminator is trained in an online updating-freezing manner where the discriminator is updated only when co-trained with the teacher generator while frozen when co-trained with the student generator. Thus, our adversarial training with an online teacher discriminator is defined as:
\begin{equation}
    \mathcal{L}_{ADV} = \mathcal{L}(D_\mathcal{T}, G_\mathcal{T}) + \mathcal{L}\big(detach(D_\mathcal{T}), G_{\mathcal{S}}\big),
\end{equation}
where $detach(\cdot)$ function detaches the gradient from the computing graph, by which, the teacher discriminator is frozen when cooperating with the student generator.
In this fashion, we demonstrate a better performance than directly discarding the discriminator~\cite{ren2021online} in Sec.\,\ref{ablation}.

\subsection{Training Loss} \label{overall}
In addition to the proposed content relationship distillation loss that considers three different content granularities and adversarial training loss with an online teacher discriminator, following~\cite{ren2021online}, we also inject image style and feature information using perceptual loss~\cite{johnson2016perceptual}. This is achieved by a pre-trained VGG-16 network~\cite{simonyan2014very} and is defined as:
\begin{equation} \label{perceptual loss}
    \begin{split}
        L_{PER} = \frac{1}{C_jH_jW_j}\Big\|\phi_j\big(f_{\mathcal{T}}(x)\big) -  \phi_j\big(f_{\mathcal{S}}(x)\big)\Big\|_1\\ 
        + \Big\|G^{\phi}_j\big(f_{\mathcal{T}}(x)\big) - G^{\phi}_j\big(f_{\mathcal{S}}(x)\big)\Big\|_1,
    \end{split}
\end{equation}
where $\phi_j(\cdot)$ returns the activation outputs with shape $C_j \times H_j \times W_j$ in the $j$-th layer of VGG network, and $G^{\phi}_j(\cdot)$ returns the $Gram$ $matrix$ of the $j$-th layer activation outputs.

Above all, the total training loss in this paper is computed as:
\begin{equation} \label{overall loss}
    \begin{split}
        L_{ALL} = L_{ADV} + \lambda_{crd} L_{CRD} + \lambda_{per} L_{PER},
    \end{split}
\end{equation}
where $\lambda_{crd}$ and $\lambda_{per}$ are two tradeoff parameters.

%------------------------------------------------------------------------
\section{Experimentation}\label{sec:experimentation}

\subsection{Experimental Settings}\label{settings}

\textbf{Models \& Datasets}.
After existing methods~\cite{jung2022exploring, zhang2022wavelet, ren2021online, li2020gan, li2022learning, li2021revisiting, jin2021teachers, shu2019co, fu2020autogan}, we perform distillation for compressed CycleGAN~\cite{zhu2017unpaired} on unpaired image-to-image translation datasets including horse2zebra and summer2winter. The former consists of 1,187 horse images and 1,474 zebra images selected from ImageNet~\cite{deng2009imagenet} while the latter contains 1,540 summer images and 1,200 winter images. We also distill compressed Pix2Pix~\cite{isola2017image} on paired image-to-image translation dataset, \emph{i.e.}, edges2shoes, which contains 50k training images from the UT Zappos50K dataset~\cite{yu2014fine} and the edge maps are computed by HED edge detection~\cite{xie2015holistically} and related post-processing methods.

\begin{table*}[!t]
    \caption{Experimental comparison on horse2zebra and summer2winter with CycleGAN. The $\Delta$ indicates performance increases over the original CycleGAN~\cite{zhu2017unpaired}.}
    \label{tab:table1}
    \vspace{-1.5em}
  \begin{center}
    \begin{tabular}{c|c|c|c|c|c}    
    \toprule
      \textbf{Dataset} & \textbf{Method} & \textbf{MACs} & \textbf{\#Parameters} & \textbf{FID($\downarrow$)} & \textbf{$\Delta$($\uparrow$)} \\
      \hline
      \multirow{12}{*}{horse2zebra} & Original~\cite{zhu2017unpaired} & 56.80G(1.0$\times$) & 11.30M(1.0$\times$) & 61.53 & - \\ 
      & Co-Evolution~\cite{shu2019co} & 13.40G(4.2$\times$) & - & 96.15 & -34.62 \\ 
      & GAN-Slimming~\cite{wang2020gan} & 11.25G(23.6$\times$) & - & 86.09 & -24.56 \\
      & AutoGAN-Distiller~\cite{fu2020autogan} & 6.39G(8.9$\times$) & - & 83.60 & -22.07 \\
      & Wavelet KD~\cite{zhang2022wavelet} & 1.68G(33.8$\times$) & 0.72M(15.81$\times$) & 77.04 & -15.51 \\
      & GAN-Compression~\cite{li2020gan} & 2.67G(21.3$\times$) & 0.34M(33.2$\times$) & 64.95 & -3.42 \\
      & Semantic Contrastive Learning~\cite{jung2022exploring} & 2.962G(19.17$\times$) & 0.41M(27.5$\times$) & 64.64 & -3.11 \\
      & DMAD~\cite{li2022learning} & 2.41G(23.6$\times$) & 0.28M(40.0$\times$) & 62.96 & -1.43 \\
      & CAT~\cite{jin2021teachers} & 2.55G(22.3$\times$) & - & 60.18 & 1.35 \\
      & GCC~\cite{li2021revisiting} & 2.40G(23.6$\times$) & - & 59.31 & 2.22 \\
      & OMGD~\cite{ren2021online} & 1.408G(40.3$\times$) & 0.137M(82.5$\times$) & 51.92 & 9.61 \\
      & \textbf{CRD} (Ours) & \textbf{1.408G(40.3$\times$)} & \textbf{0.137M(82.5$\times$)} & \textbf{46.61} & \textbf{14.92} \\ 
      \hline
      \multirow{6}{*}{summer2winter} & Original~\cite{zhu2017unpaired} & 56.80G(1.0$\times$) & 11.30M(1.0$\times$) & 79.12 & - \\
      & Co-Evolution~\cite{shu2019co} & 11.10G(5.1$\times$) & - & 78.58 & 0.54 \\
      & AutoGAN-Distiller~\cite{fu2020autogan} & 4.34G(13.1$\times$) & - & 78.33 & 0.79 \\
      & DMAD~\cite{li2022learning} & 3.18G(17.9$\times$) & 0.30M(37.7$\times$) & 78.24 & 0.88 \\
      & OMGD~\cite{ren2021online} & 1.408G(40.3$\times$) & 0.137M(82.5$\times$) & 73.79 & 5.33 \\
      & \textbf{CRD} (Ours) & \textbf{1.408G(40.3$\times$)} & \textbf{0.137M(82.5$\times$)} & \textbf{73.57} & \textbf{5.55} \\
      \bottomrule
    \end{tabular}
  \end{center}
\vspace{-1.0em}
\end{table*}

\begin{table*}[!t]
    \caption{Experimental comparison on edges2shoes with Pix2Pix. The $\Delta$ indicates the performance increases over the original Pix2Pix~\cite{isola2017image}.}
    \vspace{-1.5em}
  \begin{center}
    \begin{tabular}{c|c|c|c|c|c}    
    \toprule
      \textbf{Dataset} & \textbf{Method} & \textbf{MACs} & \textbf{\#Parameters} & \textbf{FID($\downarrow$)} & \textbf{$\Delta$($\uparrow$)} \\
      \hline
      \multirow{5}{*}{edges$\rightarrow$shoes}
      & Original~\cite{isola2017image} & 18.60G(1.0$\times$) & 54.40M(1.0$\times$) & 34.31 & - \\ 
      & Wavelet KD~\cite{zhang2022wavelet} & 1.56G(11.92$\times$) & 13.61M(4.00$\times$) & 80.13 & -45.82 \\
      & DMAD~\cite{li2022learning} & 2.99G(6.2$\times$) & 2.13M(25.5$\times$) & 46.95 & -12.64 \\ 
      & OMGD~\cite{ren2021online} & 1.219G(15.3$\times$) & 3.404M(16.0$\times$) & 25.00 & 9.41 \\
      & CRD (Ours) & \textbf{1.219G(15.3$\times$)} & \textbf{3.404M(16.0$\times$)} & \textbf{24.35} & \textbf{9.96} \\
      \bottomrule
    \end{tabular}
    \label{tab:pix2pix-edges2shoes}
  \end{center}
  \vspace{-1.5em}
\end{table*}

\textbf{Evaluation Metrics}. 
Following the previous works for GAN compression~\cite{zhang2022wavelet, ren2021online, li2021revisiting, jung2022exploring, wang2020gan, li2020gan, fu2020autogan, li2022learning, shu2019co}, we consider Fr$\Acute{\mathbf{e}}$chet Inception Distance (FID) to measure the similarity between real images and generated images. 
The algorithm is computed with the InceptionV3 model~\cite{szegedy2016rethinking} in which the last fully-connected layer is removed and feature extraction is performed. Note that a smaller FID indicates a better performance.
Besides, we also report the
MACs, parameters and their compression rates for comparison.

\textbf{Implementations}. 
The structure of our student generator is the same as the teacher generator. Following OMGD~\cite{ren2021online}, the existing SoTA method, the channel width of student generator is reduced to 1/4 of that of the teacher generator for a comparison. The input image size is fixed to 256$\times$256 and we choose 32$\times$32 as the size of our content patch (\emph{i.e.}, $n = m = $ 32).
A total of 100 training epochs are given.
%
% The learning rate is initialized to 0.0002 and linearly decayed to 0 in the end. We set batch size to 1 for CycleGAN and 4 for Pix2Pix.
The learning rate is initialized to 0.0002 and linearly decayed to 0. We set batch size to 1 for CycleGAN and 4 for Pix2Pix.
We set $\lambda_a$ and $\lambda_{per}$ to 2 and 1 respectively.
As for $\lambda_{crd}$, we set it to 25 for CycleGAN and 2.5 for Pix2Pix. More ablations regarding these hyper-parameters are provided in the appendix.
We evaluate current teacher model at an interval of 10, 6, and 1 on horse2zebra, summer2winter, and edges2shoes, and existing teachers are replaced if better-performing ones are found.% .to distill the student generator.

\subsection{Comparison}

\textbf{CycleGAN}. The quantitative experiments of compressed CycleGAN on horse2zebra and summer2winter are shown in Table\,\ref{tab:table1}. Our backbone of CycleGAN generator is based on a ResNet~\cite{he2016resnet} style to follow the previous works~\cite{li2020gan, fu2020autogan, jin2021teachers, ren2021online, gong2019autogan}. From the table, we can see that though reducing network complexity, many methods~\cite{shu2019co, wang2020gan, fu2020autogan, zhang2022wavelet, li2020gan, jung2022exploring} cause unexpected performance drops on horse2zebra dataset compared with the original CycleGAN. On the contrary, among all methods~\cite{li2022learning, jin2021teachers, li2021revisiting, ren2021online} boosting performance, our CRD leads to the best FID of 46.61 than 61.53 of the original CycleGAN with 40.3$\times$ MACs and 82.5$\times$ parameters reduction. Under the same complexity reduction, the existing SoTA OMGD~\cite{ren2021online} has 51.92 FIDs.
Similar observations can be found on summer2winter where our CRD still obtains the best performance increase of 5.55 FIDs and results in the most MACs and parameters reduction.

\textbf{Pix2Pix}. We continue the analyses on compressing Pix2Pix on edges2shoes in Table\,\ref{tab:pix2pix-edges2shoes} where two observations can be found. First, our CRD outperforms existing methods in terms of compression rates and performance increase. Second, our CRD reduces MACs and parameters of the original Pix2Pix by 15.3$\times$ and 16.0$\times$, meanwhile 9.96 FID gains are obtained.

\begin{figure*}[!t]
  \centering
  \includegraphics[width=1.0\textwidth]{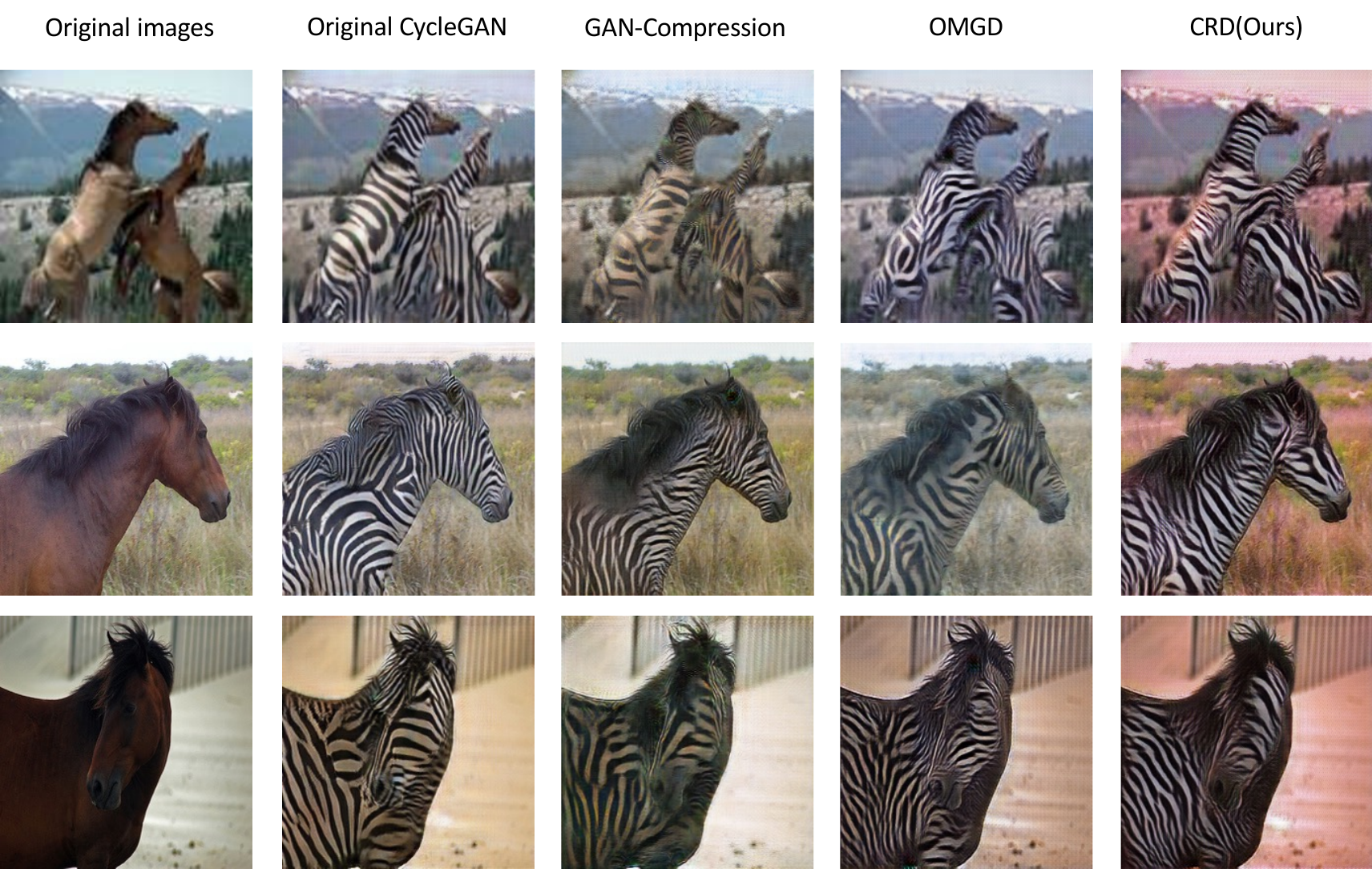}
  \vspace{-1.5em}
  \caption{Visual examples on horse2zebra dataset.}
  \label{visual_2}
  \vspace{-1.0em}
\end{figure*}

Experimental results from Table\,\ref{tab:table1} and Table\,\ref{tab:pix2pix-edges2shoes} well manifest the superiorities of our content relationship distillation in strengthening the compressed GAN models. These compressed models with better performance are in particular popular to resource-inadequate devices, which turn demonstrate the advisability of this paper to explore content relationships for distilling efficient GANs.

\textbf{Visualization}.
Lastly, we exhibit the visual image quality generated by the compressed models. Some examples of horse2zebra dataset are displayed in Fig.\,\ref{visual_2} in which we provide the original horse images and zebra images from original CycleGAN~\cite{zhu2017unpaired}, typical GAN-Compression~\cite{li2020gan}, the recent SoTA OMGD~\cite{ren2021online} as well as our proposed CRD. All zebra images are produced from the compressed models presented in Table\,\ref{tab:table1}. The proposed CRD not only advantages in higher model complexity reduction, but also outstands in its vivid visual effect. Our CRD is prone to resulting in stripes of white and black while existing compressed methods produce gold-black styles. Even compared to the original CycleGAN, colors in the proposed method are more clear. In addition, our zebra texture generation is very smooth while other methods cause chaos in the region between the ``neck'' and ``body''.

\subsection{Ablation Study}\label{ablation}

We take CycleGAN on horse2zebra as an example to conduct ablation studies, including our content-level relationship distillation \emph{v.s.} instance-level counterpart~\cite{park2019relational}, the content granularity, patch sizes in our image patch granularity and our online teacher discriminator. The results are provided from Table\,\ref{tab:ablation different granularity} to Table\,\ref{tab:discrimnator} where the ``Baseline'' means training the student model without distillation.

\begin{table}[!t]
 \centering
 \caption{Comparison between instance-level relationship distillation~\cite{park2019relational} and our content-level relationship distillation.}
  \vspace{-0.5em}
 \begin{tabular}{c|c|c}\toprule
    \textbf{Dataset} & \textbf{Method} & \textbf{FID($\downarrow$)} \\
    \hline
    \multirow{4}{*}{horse2zebra} 
    & Baseline (w/o KD) & 96.72 \\
    & Instance level (batch size 4) & 63.93 \\
    & Instance level (batch size 8) & 67.98 \\
    \cline{2-3}
    & Ours & 46.61 \\
    \bottomrule
 \end{tabular}
 \label{tab:ablation different granularity}
 \vspace{-1.0em}
\end{table}

\begin{table}[!t]
 \centering
  \caption{Ablations of content granularities including image column, row and patch.}
  \vspace{-0.5em}
 \begin{tabular}{c|c|c}\toprule
    \textbf{Dataset} & \textbf{Method} & \textbf{FID($\downarrow$)} \\
    \hline
    \multirow{5}{*}{horse2zebra} 
    & Baseline (w/o KD) & 96.72 \\
    & Ours w/o Patch & 49.08 \\
    & Ours w/o Row & 55.35 \\
    & Ours w/o Column & 48.93 \\
    \cline{2-3}
    & Ours & 46.61 \\
    \bottomrule
 \end{tabular}
 \label{tab:ablation same granularity}
 \vspace{-1.5em}
\end{table}

\textbf{Relationship Level}.
Our CRD explores the content-level relationships in an image. In contrast, the vanilla relational knowledge distillation~\cite{park2019relational} excavates instance-level relationships among images. Here, we replace our content-level relationship distillation with instance-level relationship distillation of different training batch sizes, and show the comparison in Table\,\ref{tab:ablation different granularity}.
Results with instance-level relationship distillation do improve over the vanilla Baseline without any distillation from 96.72 to 63.93 with batch size 4 and 67.98 with batch size 8. Nevertheless, the improvement is quite limited if compared to our CRD of 46.61 FIDs. These results demonstrate our claim in Sec.\,\ref{Patch-wise Relational Knowledge Distillation} that generative tasks focus more on image contents and content-level relationships are more favourable in GAN compression.

\textbf{Content Granularity}. 
Table\,\ref{tab:ablation same granularity} offers ablations \emph{w.r.t}. our three content granularities. We verify their effectiveness by individually removing one of them. Two observations from Table\,\ref{tab:ablation same granularity} can be found. First, removing any content granularity drops the performance, among which, the row granularity causes the most FID decrease. This exemplifies the importance of our three content granularities. Second, despite the performance decrease, removing a single content granularity still leads to performance increase in comparison with the Baseline. This inspires us of the correctness of excavating content relationships in distilling GANs.

\begin{table}[!t]
 \centering
  \caption{Ablations of patch sizes in content patch relationships.}
  \vspace{-0.5em}
 \begin{tabular}{c|c|c}\toprule
    \textbf{Dataset} & \textbf{Method} & \textbf{FID($\downarrow$)} \\
    \hline
    \multirow{4}{*}{horse2zebra} 
    & Baseline (w/o KD) & 96.72 \\
    & Patch size 16$\times$16 & 55.39 \\
    & Patch size 64$\times$64 & 60.91 \\
    \cline{2-3}
    & Ours (patch size 32$\times$32) & 46.61 \\
    \bottomrule
 \end{tabular}
 \label{tab:ablation different size of patch}
 \vspace{-1.5em}
\end{table}

\begin{figure}[!t]
  \centering
  \includegraphics[width=0.5\textwidth]{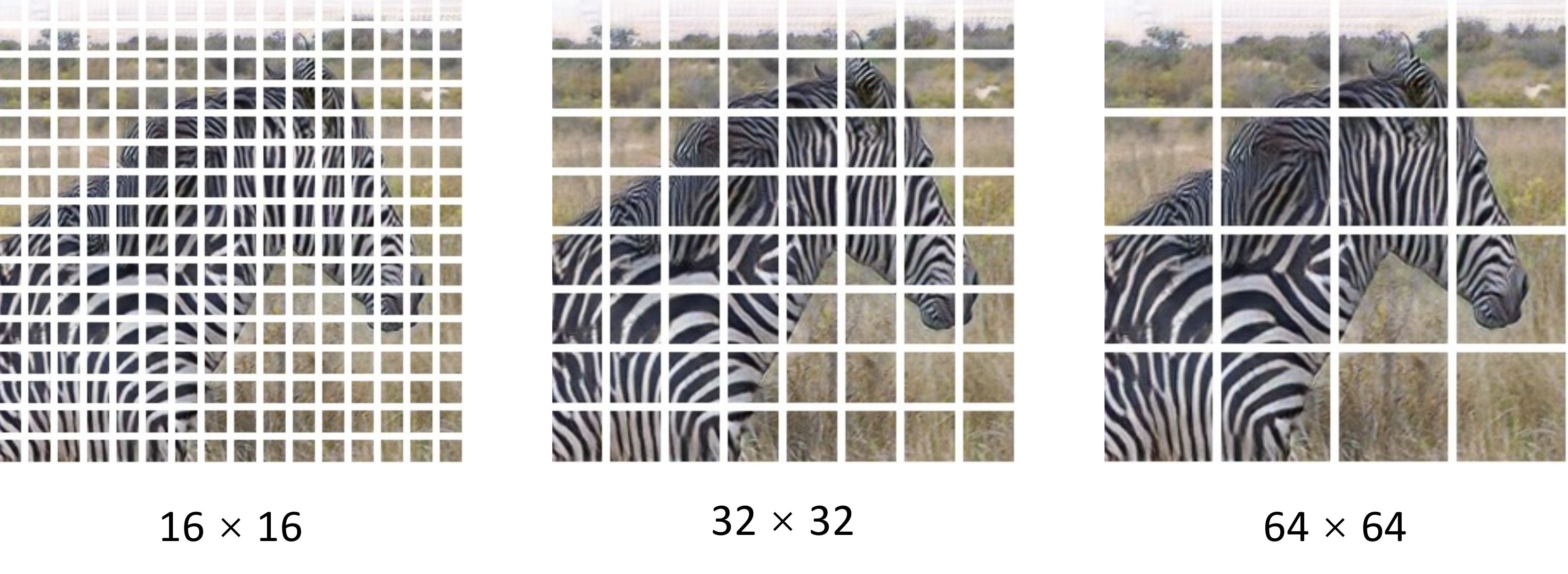}
  \vspace{-1.5em}
  \caption{Splitting a 256$\times$256 image into patches of 16$\times$16, 32$\times$32 and 64$\times$64.}
  \label{content patch size}
\end{figure}

\textbf{Content Patch Size}. 
We set the patch size in our patch granularity as 32$\times$32 as stated in Sec.\,\ref{settings}. Herein, we further enlarge it to 64$\times$64 and reduce it to 16$\times$16 to manifest the effect. Results in Table\,\ref{tab:ablation different size of patch} show that our CRD is sensitive to the patch sizes and 32$\times$32 provides the best option. As shown in Fig.\,\ref{content patch size}, the possible reason is attributed to the explanation as follows.
Given an image of 256$\times$256, a total of 256 non-overlapping patches can be obtained for 16$\times$16 patch size, each of which is too small and therefore destroys the locality.  On the contrary, a 64$\times$64 patch is too large, which tends to embrace more global information and contradicts our row/column stripes that also focus on global information. Our 32$\times$32 patch is just right to model the local information and is complementary to the global row/column stripes to pursue better performance.

\textbf{Online Teacher Discriminator}.
%
%We lastly knuckle down to the efficacy of our proposed online teacher discriminator. 
Recall that in Sec.\,\ref{online nash}, we analyze that a pre-trained teacher discriminator is too strong for our from-scratch-training generator and crashes Nash equilibrium in adversarial training. Accordingly, we introduce an online teacher discriminator trained in an updating-freezing manner. To verify our analyses, we replace our online teacher discriminator with a well pre-trained one which is updated or frozen during training. Also, we remove our discriminator when co-trained with student generator following~\cite{ren2021online}. 
Table\,\ref{tab:discrimnator} presents the experimental performance. We can see that a pre-trained teacher discriminator in any case of freezing or updating, suffers poor performance compared with an online teacher discriminator which manifests greater performance of 62.75 even in the worst case of no discriminator when co-trained with student generator.
%
% Therefore, an online teacher discriminator is more suitable for our content relationship distillation. Also, when substituting our updating-freezing manner with a simple updating fashion where the discriminator is updated across co-trained with both teacher and student generators, FID drops from 46.61 to 58.02. When adopting the scenario from~\cite{ren2021online} that removes teacher discriminator when co-trained with student generator, the performance drops to 62.75.
%
Therefore, an online teacher discriminator is more suitable for our content relationship distillation. Also, when substituting our updating-freezing manner with a simple updating fashion where the discriminator is updated across co-trained with both teacher and student generators, FID drops from 46.61 to 58.02. When adopting the scenario from~\cite{ren2021online} which removes discriminator when training student generator, the performance drops to 62.75.
Thus, our updating-freezing training benefits more from the adversarial training in GAN compression.

\begin{table}[!t]
 \centering
  \caption{Ablations of our online teacher discriminator.}
  \vspace{-0.5em}
 \begin{tabular}{c|c|c}\toprule
    \textbf{Dataset} & \textbf{Method} & \textbf{FID($\downarrow$)} \\
    \hline
    \multirow{4}{*}{horse2zebra} 
    & Pre-trained (freezing) & 69.48 \\
    & Pre-trained (updating) & 70.41 \\
%    & Pre-trained (updating-freezing) & \\
    \cline{2-3}
    & Online (w/o discriminator) & 62.75 \\
    & Online (always updating) & 58.02 \\
    & Our online (updating-freezing) & 46.61 \\ 
    \bottomrule
 \end{tabular}
 \label{tab:discrimnator}
 \vspace{-1.0em}
\end{table}

\section{Future Study}\label{sec:limitations}
%
% Though great advancement has been made in this paper, we would like to raise our concerns \emph{w.r.t.} the background color of our visual results. Diving into a deep observation in Fig.\,\ref{visual_2}, we can find a better visual quality of the synthesized zebra object, however, the background color is damaged if compared to the original images. Though this issue occurs to other methods as well, more efforts will be made in our subsequent work to solve it by possibly aligning the color channels or adding a priori channel knowledge during training networks.
Though great advancement has been made in this paper, we would like to raise our concerns \emph{w.r.t.} the background color of our visual results. Diving into a deep observation in Fig.\,\ref{visual_2}, we can find a better visual quality of the synthesized zebra object, however, the background color is damaged compared to the original images. Though this issue occurs to other methods as well, more efforts will be made in our subsequent work to solve it by possibly aligning the color channels or adding a priori channel knowledge.

\section{Conclusion}\label{sec:conclusion}
We presented a content relationship distillation method (CRD) in this paper to strengthen the performance of compressed GANs. Our CRD models content-level relationships for the first time to provide GANs oriented knowledge. The contents include row/column stripes for extracting global information and image patches for extracting local information. Also, an online teacher discriminator is proposed to online co-trained with the teacher and student generators for better adversarial training.
We have conducted extensive experiments, which show that our CRD reaches the most reduction in GANs complexity while obtaining the best performance. Ablations also demonstrate the efficacy of each part in our method.

\section*{Acknowledgement}
This work is supported by the National Science Fund for Distinguished Young (No.62025603), the National Natural Science Foundation of China (No.62025603, No. U1705262, No. 62072386, No. 62072387, No. 62072389, No, 62002305, No.61772443, No. 61802324 and No. 61702136) and Guangdong Basic and Applied Basic Research Foundation (No.2019B1515120049).

%%%%%%%%% REFERENCES
{\small
\bibliographystyle{ieee_fullname}
\bibliography{main}

\begin{thebibliography}{10}\itemsep=-1pt

\bibitem{aguinaldo2019compressing}
Angeline Aguinaldo, Ping-Yeh Chiang, Alex Gain, Ameya Patil, Kolten Pearson,
  and Soheil Feizi.
\newblock Compressing gans using knowledge distillation.
\newblock {\em arXiv preprint arXiv:1902.00159}, 2019.

\bibitem{brock2018large}
Andrew Brock, Jeff Donahue, and Karen Simonyan.
\newblock Large scale gan training for high fidelity natural image synthesis.
\newblock {\em arXiv preprint arXiv:1809.11096}, 2018.

\bibitem{chen2020distilling}
Hanting Chen, Yunhe Wang, Han Shu, Changyuan Wen, Chunjing Xu, Boxin Shi, Chao
  Xu, and Chang Xu.
\newblock Distilling portable generative adversarial networks for image
  translation.
\newblock In {\em Proceedings of the AAAI Conference on Artificial Intelligence
  (AAAI)}, pages 3585--3592, 2020.

\bibitem{chen2021wasserstein}
Liqun Chen, Dong Wang, Zhe Gan, Jingjing Liu, Ricardo Henao, and Lawrence
  Carin.
\newblock Wasserstein contrastive representation distillation.
\newblock In {\em Proceedings of the IEEE/CVF Conference on Computer Vision and
  Pattern Recognition (CVPR)}, pages 16296--16305, 2021.

\bibitem{chen2021distilling}
Pengguang Chen, Shu Liu, Hengshuang Zhao, and Jiaya Jia.
\newblock Distilling knowledge via knowledge review.
\newblock In {\em Proceedings of the IEEE/CVF Conference on Computer Vision and
  Pattern Recognition (CVPR)}, pages 5008--5017, 2021.

\bibitem{chen2021gans}
Xuxi Chen, Zhenyu Zhang, Yongduo Sui, and Tianlong Chen.
\newblock Gans can play lottery tickets too.
\newblock {\em arXiv preprint arXiv:2106.00134}, 2021.

\bibitem{chen2018cartoongan}
Yang Chen, Yu-Kun Lai, and Yong-Jin Liu.
\newblock Cartoongan: Generative adversarial networks for photo cartoonization.
\newblock In {\em Proceedings of the IEEE/CVF Conference on Computer Vision and
  Pattern Recognition (CVPR)}, pages 9465--9474, 2018.

\bibitem{choi2018stargan}
Yunjey Choi, Minje Choi, Munyoung Kim, Jung-Woo Ha, Sunghun Kim, and Jaegul
  Choo.
\newblock Stargan: Unified generative adversarial networks for multi-domain
  image-to-image translation.
\newblock In {\em Proceedings of the IEEE/CVF Conference on Computer Vision and
  Pattern Recognition (CVPR)}, pages 8789--8797, 2018.

\bibitem{deng2009imagenet}
Jia Deng, Wei Dong, Richard Socher, Li-Jia Li, Kai Li, and Li Fei-Fei.
\newblock Imagenet: A large-scale hierarchical image database.
\newblock In {\em Proceedings of the IEEE/CVF Conference on Computer Vision and
  Pattern Recognition (CVPR)}, pages 248--255, 2009.

\bibitem{fu2020autogan}
Yonggan Fu, Wuyang Chen, Haotao Wang, Haoran Li, Yingyan Lin, and Zhangyang
  Wang.
\newblock Autogan-distiller: Searching to compress generative adversarial
  networks.
\newblock {\em arXiv preprint arXiv:2006.08198}, 2020.

\bibitem{gatys2015texture}
Leon Gatys, Alexander~S Ecker, and Matthias Bethge.
\newblock Texture synthesis using convolutional neural networks.
\newblock {\em Advances in Neural Information Processing Systems (NeurIPS)},
  2015.

\bibitem{gatys2016image}
Leon~A Gatys, Alexander~S Ecker, and Matthias Bethge.
\newblock Image style transfer using convolutional neural networks.
\newblock In {\em Proceedings of the IEEE/CVF Conference on Computer Vision and
  Pattern Recognition (CVPR)}, pages 2414--2423, 2016.

\bibitem{gong2019autogan}
Xinyu Gong, Chang Shiyu, Yifan Jiang, and Zhangyang Wang.
\newblock Autogan: Neural architecture search for generative adversarial
  networks.
\newblock In {\em Proceedings of the IEEE/CVF Conference on Computer Vision and
  Pattern Recognition (CVPR)}, pages 3224--3234, 2019.

\bibitem{goodfellow2014generative}
Ian Goodfellow, Jean Pouget-Abadie, Mehdi Mirza, Bing Xu, David Warde-Farley,
  Sherjil Ozair, Aaron Courville, and Yoshua Bengio.
\newblock Generative adversarial nets.
\newblock {\em Advances in Neural Information Processing Systems (NeurIPS)},
  2014.

\bibitem{han2015deep}
Song Han, Huizi Mao, and William~J Dally.
\newblock Deep compression: Compressing deep neural networks with pruning,
  trained quantization and huffman coding.
\newblock {\em arXiv preprint arXiv:1510.00149}, 2015.

\bibitem{he2016resnet}
Kaiming He, Zhang Xiangyu, Ren Shaoqing, and Jian Sun.
\newblock Deep residual learning for image recognition.
\newblock In {\em Proceedings of the IEEE/CVF Conference on Computer Vision and
  Pattern Recognition (CVPR)}, pages 770--778, 2016.

\bibitem{heo2019comprehensive}
Byeongho Heo, Jeesoo Kim, Sangdoo Yun, Hyojin Park, Nojun Kwak, and Jin~Young
  Choi.
\newblock A comprehensive overhaul of feature distillation.
\newblock In {\em Proceedings of the IEEE/CVF International Conference on
  Computer Vision (ICCV)}, pages 1921--1930, 2019.

\bibitem{hinton2015distilling}
Geoffrey Hinton, Oriol Vinyals, Jeff Dean, et~al.
\newblock Distilling the knowledge in a neural network.
\newblock {\em arXiv preprint arXiv:1503.02531}, 2015.

\bibitem{hou2021slimmable}
Liang Hou, Zehuan Yuan, Lei Huang, Huawei Shen, Xueqi Cheng, and Changhu Wang.
\newblock Slimmable generative adversarial networks.
\newblock In {\em Proceedings of the AAAI Conference on Artificial Intelligence
  (AAAI)}, pages 7746--7753, 2021.

\bibitem{isola2017image}
Phillip Isola, Jun-Yan Zhu, Tinghui Zhou, and Alexei~A Efros.
\newblock Image-to-image translation with conditional adversarial networks.
\newblock In {\em Proceedings of the IEEE/CVF Conference on Computer Vision and
  Pattern Recognition (CVPR)}, pages 1125--1134, 2017.

\bibitem{yu2018slimmable}
Yu Jiahui, Yang Linjie, Xu Ning, Yang Jianchao, and Huang Thomas.
\newblock Slimmable neural networks.
\newblock {\em arXiv preprint arXiv:1812.08928}, 2018.

\bibitem{jin2021teachers}
Qing Jin, Jian Ren, Oliver~J Woodford, Jiazhuo Wang, Geng Yuan, Yanzhi Wang,
  and Sergey Tulyakov.
\newblock Teachers do more than teach: Compressing image-to-image models.
\newblock In {\em Proceedings of the IEEE/CVF Conference on Computer Vision and
  Pattern Recognition (CVPR)}, pages 13600--13611, 2021.

\bibitem{johnson2016perceptual}
Justin Johnson, Alahi Alexandre, and Fei-Fei Li.
\newblock Perceptual losses for real-time style transfer and super-resolution.
\newblock In {\em European Conference on Computer Vision (ECCV)}, pages
  694--711, 2016.

\bibitem{jung2022exploring}
Chanyong Jung, Gihyun Kwon, and Jong~Chul Ye.
\newblock Exploring patch-wise semantic relation for contrastive learning in
  image-to-image translation tasks.
\newblock In {\em Proceedings of the IEEE/CVF Conference on Computer Vision and
  Pattern Recognition (CVPR)}, pages 18260--18269, 2022.

\bibitem{karras2019style}
Tero Karras, Samuli Laine, and Timo Aila.
\newblock A style-based generator architecture for generative adversarial
  networks.
\newblock In {\em Proceedings of the IEEE/CVF Conference on Computer Vision and
  Pattern Recognition (CVPR)}, pages 4401--4410, 2019.

\bibitem{karras2020analyzing}
Tero Karras, Samuli Laine, Miika Aittala, Janne Hellsten, Jaakko Lehtinen, and
  Timo Aila.
\newblock Analyzing and improving the image quality of stylegan.
\newblock In {\em Proceedings of the IEEE/CVF Conference on Computer Vision and
  Pattern Recognition (CVPR)}, pages 8110--8119, 2020.

\bibitem{li2020gan}
Muyang Li, Ji Lin, Yaoyao Ding, Zhijian Liu, Jun-Yan Zhu, and Song Han.
\newblock Gan compression: Efficient architectures for interactive conditional
  gans.
\newblock In {\em Proceedings of the IEEE/CVF Conference on Computer Vision and
  Pattern Recognition (CVPR)}, pages 5284--5294, 2020.

\bibitem{li2022learning}
Shaojie Li, Mingbao Lin, Yan Wang, Chao Fei, Ling Shao, and Rongrong Ji.
\newblock Learning efficient gans for image translation via differentiable
  masks and co-attention distillation.
\newblock In {\em IEEE Transactions on Multimedia (TMM)}, 2022.

\bibitem{li2021revisiting}
Shaojie Li, Jie Wu, Xuefeng Xiao, Fei Chao, Xudong Mao, and Rongrong Ji.
\newblock Revisiting discriminator in gan compression: A
  generator-discriminator cooperative compression scheme.
\newblock {\em Advances in Neural Information Processing Systems (NeurIPS)},
  pages 28560--28572, 2021.

\bibitem{li2020semantic}
Zeqi Li, Ruowei Jiang, and Parham Aarabi.
\newblock Semantic relation preserving knowledge distillation for
  image-to-image translation.
\newblock In {\em European Conference on Computer Vision (ECCV)}, pages
  648--663, 2020.

\bibitem{lin2021anycost}
Ji Lin, Richard Zhang, Frieder Ganz, Song Han, and Jun-Yan Zhu.
\newblock Anycost gans for interactive image synthesis and editing.
\newblock In {\em Proceedings of the IEEE/CVF Conference on Computer Vision and
  Pattern Recognition (CVPR)}, pages 14986--14996, 2021.

\bibitem{nash1951non}
John Nash.
\newblock Non-cooperative games.
\newblock {\em Annals of mathematics}, pages 286--295, 1951.

\bibitem{park2019relational}
Wonpyo Park, Dongju Kim, Yan Lu, and Minsu Cho.
\newblock Relational knowledge distillation.
\newblock In {\em Proceedings of the IEEE/CVF Conference on Computer Vision and
  Pattern Recognition (CVPR)}, pages 3967--3976, 2019.

\bibitem{radford2015unsupervised}
Alec Radford, Luke Metz, and Soumith Chintala.
\newblock Unsupervised representation learning with deep convolutional
  generative adversarial networks.
\newblock {\em arXiv preprint arXiv:1511.06434}, 2015.

\bibitem{ren2021online}
Yuxi Ren, Jie Wu, Xuefeng Xiao, and Jianchao Yang.
\newblock Online multi-granularity distillation for gan compression.
\newblock In {\em Proceedings of the IEEE/CVF International Conference on
  Computer Vision (ICCV)}, pages 6793--6803, 2021.

\bibitem{romero2014fitnets}
Adriana Romero, Nicolas Ballas, Samira~Ebrahimi Kahou, Antoine Chassang, Carlo
  Gatta, and Yoshua Bengio.
\newblock Fitnets: Hints for thin deep nets.
\newblock {\em arXiv preprint arXiv:1412.6550}, 2014.

\bibitem{shu2019co}
Han Shu, Yunhe Wang, Xu Jia, Kai Han, Hanting Chen, Chunjing Xu, Qi Tian, and
  Chang Xu.
\newblock Co-evolutionary compression for unpaired image translation.
\newblock In {\em Proceedings of the IEEE/CVF International Conference on
  Computer Vision (ICCV)}, pages 3235--3244, 2019.

\bibitem{simonyan2014very}
Karen Simonyan and Andrew Zisserman.
\newblock Very deep convolutional networks for large-scale image recognition.
\newblock {\em arXiv preprint arXiv:1409.1556}, 2014.

\bibitem{szegedy2016rethinking}
Christian Szegedy, Vincent Vanhoucke, Sergey Ioffe, Jon Shlens, and Zbigniew
  Wojna.
\newblock Rethinking the inception architecture for computer vision.
\newblock In {\em Proceedings of the IEEE/CVF Conference on Computer Vision and
  Pattern Recognition (CVPR)}, pages 2818--2826, 2016.

\bibitem{tian2019contrastive}
Yonglong Tian, Dilip Krishnan, and Phillip Isola.
\newblock Contrastive representation distillation.
\newblock {\em arXiv preprint arXiv:1910.10699}, 2019.

\bibitem{wang2020gan}
Haotao Wang, Shupeng Gui, Haichuan Yang, Ji Liu, and Zhangyang Wang.
\newblock Gan slimming: All-in-one gan compression by a unified optimization
  framework.
\newblock In {\em European Conference on Computer Vision (ECCV)}, pages 54--73,
  2020.

\bibitem{wang2019qgan}
Peiqi Wang, Dongsheng Wang, Yu Ji, Xinfeng Xie, Haoxuan Song, XuXin Liu,
  Yongqiang Lyu, and Yuan Xie.
\newblock Qgan: Quantized generative adversarial networks.
\newblock {\em arXiv preprint arXiv:1901.08263}, 2019.

\bibitem{wang2018kdgan}
Xiaojie Wang, Rui Zhang, Yu Sun, and Jianzhong Qi.
\newblock Kdgan: Knowledge distillation with generative adversarial networks.
\newblock {\em Advances in Neural Information Processing Systems (NeurIPS)},
  2018.

\bibitem{xie2015holistically}
Saining Xie and Tu Zhuowen.
\newblock Holistically-nested edge detection.
\newblock In {\em Proceedings of the IEEE/CVF Conference on Computer Vision and
  Pattern Recognition (CVPR)}, pages 1395--1403, 2015.

\bibitem{xu2021drb}
Wenju Xu, Chengjiang Long, Ruisheng Wang, and Guanghui Wang.
\newblock Drb-gan: A dynamic resblock generative adversarial network for
  artistic style transfer.
\newblock In {\em Proceedings of the IEEE/CVF International Conference on
  Computer Vision (ICCV)}, pages 6383--6392, 2021.

\bibitem{yu2014fine}
Aron Yu and Grauman Kristen.
\newblock Fine-grained visual comparisons with local learning.
\newblock In {\em Proceedings of the IEEE/CVF Conference on Computer Vision and
  Pattern Recognition (CVPR)}, pages 192--199, 2014.

\bibitem{zhang2019self}
Han Zhang, Ian Goodfellow, Dimitris Metaxas, and Augustus Odena.
\newblock Self-attention generative adversarial networks.
\newblock In {\em International Conference on Machine Learning (ICML)}, pages
  7354--7363, 2019.

\bibitem{zhang2022wavelet}
Linfeng Zhang, Xin Chen, Xiaobing Tu, Pengfei Wan, Ning Xu, and Kaisheng Ma.
\newblock Wavelet knowledge distillation: Towards efficient image-to-image
  translation.
\newblock In {\em Proceedings of the IEEE/CVF Conference on Computer Vision and
  Pattern Recognition (CVPR)}, pages 12464--12474, 2022.

\bibitem{zhao2022decoupled}
Borui Zhao, Quan Cui, Renjie Song, Yiyu Qiu, and Jiajun Liang.
\newblock Decoupled knowledge distillation.
\newblock In {\em Proceedings of the IEEE/CVF Conference on Computer Vision and
  Pattern Recognition (CVPR)}, pages 11953--11962, 2022.

\bibitem{zhu2017unpaired}
Jun-Yan Zhu, Taesung Park, Phillip Isola, and Alexei~A Efros.
\newblock Unpaired image-to-image translation using cycle-consistent
  adversarial networks.
\newblock In {\em Proceedings of the IEEE International Conference on Computer
  Vision (ICCV)}, pages 2223--2232, 2017.

\end{thebibliography}
}

\clearpage

\section*{Appendix \label{appendix}}

\section{Additional Ablation Studies}

\begin{table}[htbp]
 \centering
  \caption{Ablation studies of $\lambda_{crd}$.}
 \label{tab:lambda_crd}
 \begin{tabular}{c|c|c}\toprule
    \textbf{Dataset} & \textbf{$\lambda_{crd}$} & \textbf{FID($\downarrow$)} \\
    \hline
    \multirow{3}{*}{horse2zebra} 
    & 10 & 68.25 \\
    & 50 & 58.11 \\
    \cline{2-3}
    & Ours (25) & 46.61 \\
    \bottomrule
 \end{tabular}
\end{table}

\textbf{Trade-off parameter of content relationship distillation}. Table\,\ref{tab:lambda_crd} shows the ablation study on content relationship distillation hyper-parameter $\lambda_{crd}$. In the table, the performance of the compressed GANs is significantly improved when using CRD loss, i.e. $\lambda_{crd} = 25$, and too small value of CRD loss cannot get enough content relationship which leads to poor performance.

\begin{table}[htbp]
 \centering
  \caption{Ablation studies of $\lambda_a$.}
 \label{tab:lambda_a}
 \begin{tabular}{c|c|c}\toprule
    \textbf{Dataset} & \textbf{$\lambda_{a}$} & \textbf{FID($\downarrow$)} \\
    \hline
    \multirow{3}{*}{horse2zebra} 
    & 0.5 & 58.67 \\
    & 1 & 52.63 \\
    \cline{2-3}
    & Ours (2) & 46.61 \\
    \bottomrule
 \end{tabular}
\end{table}

\textbf{Trade-off parameter between distance and angle}. Table\,\ref{tab:lambda_a} shows ablation study of the balance parameter $\lambda_{a}$, which is measurement of the distance and angle. In the experimental results, it is shown that the most primitive hyper-parameter in the traditional relational distillation works best without additional changes.

\begin{table}[htbp]
 \centering
  \caption{Ablation studies of $\lambda_{per}$.}
 \label{tab:lambda_per}
 \begin{tabular}{c|c|c}\toprule
    \textbf{Dataset} & \textbf{$\lambda_{per}$} & \textbf{FID($\downarrow$)} \\
    \hline
    \multirow{3}{*}{horse2zebra} 
    & 2 & 57.60 \\
    & 3 & 58.35 \\
    \cline{2-3}
    & Ours (1) & 46.61 \\
    \bottomrule
 \end{tabular}
\end{table}

\textbf{Trade-off parameter of perceptual loss}. Table\,\ref{tab:lambda_per} shows the balance parameter with perceptual loss $\lambda_{per}$ , which extracts the characterization information of style and feature, and we intuitively set the hyper-parameter of perceptual loss to 1 to achieve the optimal effect.

\section{Additional Comparison}

\begin{figure*}[!t]
  \centering
  \includegraphics[width=1.0\textwidth]{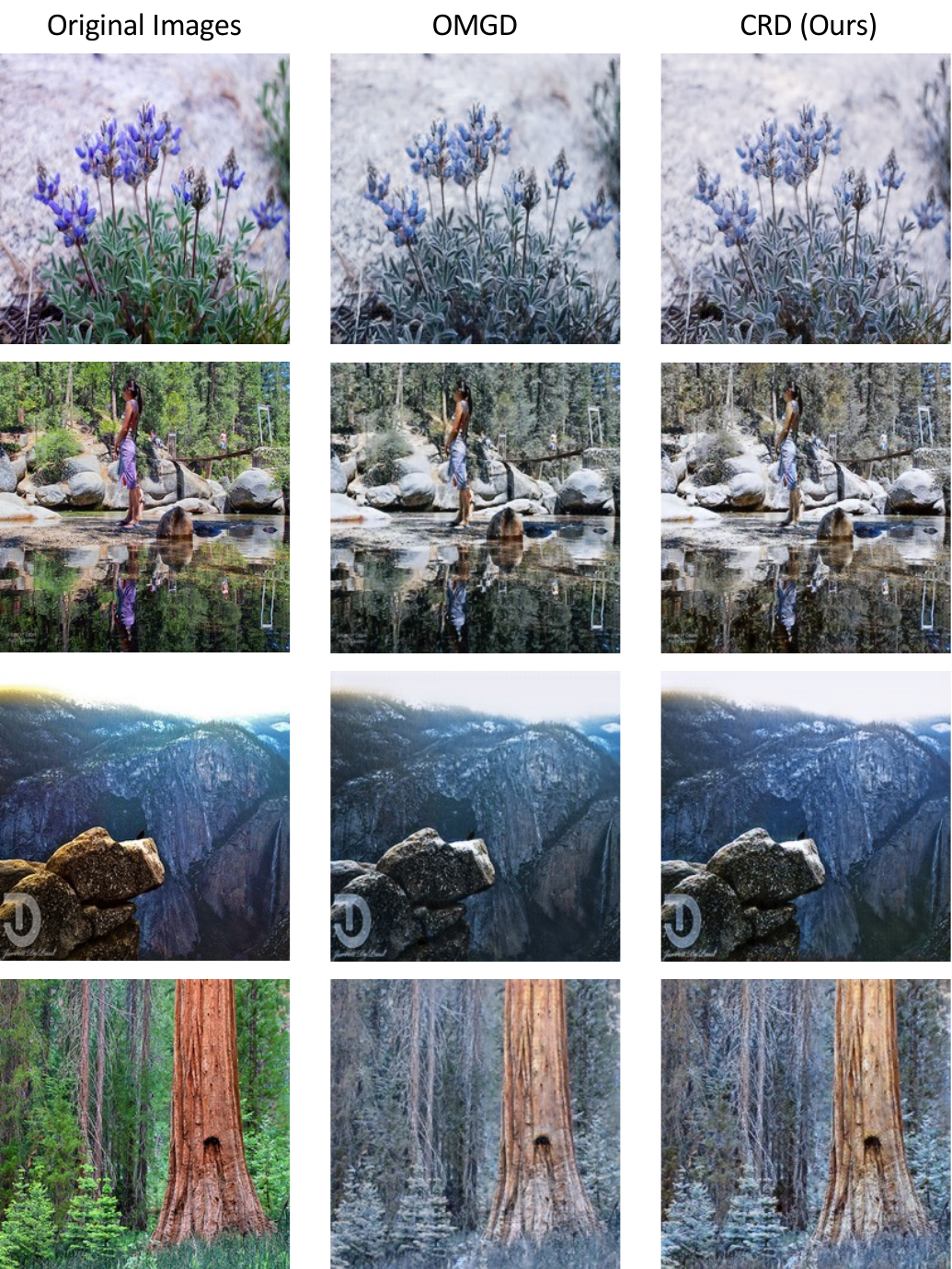}
  \vspace{-1.5em}
  \caption{Visual examples on summer2winter dataset.}
  \label{summer2winter}
  \vspace{-1.0em}
\end{figure*}

\textbf{Visualization}. We have additionally added a visual presentation of compressing CycleGAN on the summer2winter dataset here. In Fig.\,\ref{summer2winter}, we show the input images, the OMGD results (current SoTA method), and our CRD results. The visual results demonstrate that by simply transforming the season in summer images, the visual effects of our method are still more clear. Compared with the OMGD method, our CRD results are more natural.

\end{document}